\documentclass{article} 
\usepackage{iclr2015,times}
\usepackage{hyperref}
\usepackage{url}
\usepackage{amsmath}
\usepackage{amssymb}
\usepackage{amsthm}
\usepackage{subcaption}
\usepackage{todonotes}
\usepackage{pbox}
\usepackage{stmaryrd}

\title{On Learning Vector Representations \\in Hierarchical Label Spaces}

\author{
Jinseok Nam\textsuperscript{1,2}, Johannes F{\"u}rnkranz\textsuperscript{1} \\
\textsuperscript{1} Knowledge Engineering Group\\
Department of Computer Science\\
Technische Universit{\"a}t Darmstadt \vspace{3pt}\\
\textsuperscript{2} Knowledge Discovery in Scientific Literature\\
German Institute for Educational Research\\
\texttt{\{nam@cs, juffi@ke\}.tu-darmstadt.de}
}

\iclrfinalcopy 

\begin{document}

\maketitle

\begin{abstract}
An important problem in multi-label classification is to 
capture label patterns or underlying structures that have an impact on such patterns.
This paper addresses one such problem, namely
how to exploit hierarchical structures over labels.
We present a novel method to learn vector representations of a label space given a hierarchy of labels and label co-occurrence patterns.
Our experimental results demonstrate qualitatively that the proposed method is able to learn regularities among labels by exploiting a label hierarchy as well as label co-occurrences.
It highlights the importance of the hierarchical information in order to obtain regularities which facilitate analogical reasoning over a label space. 
We also experimentally illustrate the dependency of the learned representations on the label hierarchy.
\end{abstract}

\section{Introduction} \label{sec:intro}
Multi-label classification is an area of machine learning which aims to learn a function that maps instances to label spaces.
In contrast to multi-class classification, each instance is assumed to be associated with more than one label.
One of the goals in multi-label classification is to model underlying structures over a label space because in many such problems, 
the occurrence of labels is not independent of each other.
Many attempts have been made to capture and exploit label structures \citep{Fuernkranz2008,hsu2009multi,read2011classifier,dembczynski2012label}.
As the number of possible configurations of labels grows exponentially with respect to the number of labels, it is required for multi-label classifier to handle many labels efficiently \citep{bi2013efficient}.
Not only multi-label classifiers but also human annotators need a way to handle a large number of labels efficiently. 
Thus, human experts put a lot of effort into maintaining and updating hierarchies of labels and such hierarchies are used to generate the ground truth for training classifiers, and many methods have been developed to use hierarchical output structures in machine learning \citep{HierarchicalClassification-survey}.
In particular, several researchers have looked into utilizing the hierarchical structure of the label space for improved predictions in multi-label classification \citep{multilabel:HierarchicalKernel,multilabel:hierarchical-dts,multilabel:flat,bi2011multi}.

In this work, we present a novel method to efficiently learn from hierarchical structures over labels as well as occurrences of labels, and investigate the importance of hierarchical structures to identify internal structures of labels.

The rest of the paper is organized as follows.
In Section \ref{sec:problem_definition}, we define multi-label classification and hierarchical structures of labels in a graph.
We then introduce our proposed method that learns label spaces while taking into account label hierarchies.
In Section \ref{sec:exp_setup}, we set up our experiments, and present empirical analysis in Section \ref{sec:exp_results}.
Finally, we discuss findings from our experiments and provide directions for future work in Section \ref{sec:discussion}.

\section{Multi-label Classification with Label Hierarchies} \label{sec:problem_definition}
Throughout this work we will present a method to learn representations of labels in multi-label classification. Firstly, in this section we define multi-label classification and notation for label hierarchies.

{\bf Notation. }Multi-label classification refers to the task of learning a function that maps instances $\mathbf{x}$ to label sets $\mathcal{Y} \subseteq \{1,2,\cdots,L\}$ given a training dataset $\mathcal{D} = \{(\mathbf{x}_n, \mathcal{Y}_n)\}_{n=1}^{N}$, where $N$ is the number of instances and $L$ is the number of labels.
Consider multi-label problems where label hierarchies exist.
Label graphs are a natural way to represent such hierarchical structures.
Because it is possible for a label to have more than one parent node, we represent a hierarchy of labels in a directed acyclic graph (DAG).
Consider a graph $\mathcal{G} = \{V, E\}$ where $V$ denotes a node and $E$ represent a connection between two nodes.
Each node corresponds a label and an edge represents 
a parent-child relationship between the two labels in the hierarchy.
A node $u \in V$ corresponds to a label.
If there exists a directed path from $u$ to $v$, $u$ is an ancestor of $v$ and $v$ is a descendant of $u$.
The set of all ancestors of $v$ is denoted as $\mathcal{S}_{\mathbf{A}}(v)$, and the set of all descendants is denoted as $\mathcal{S}_{\mathbf{D}}(u)$.

For large-scale multi-label classification problems, high dimensionality of label spaces makes learning classifiers difficult.
In the literature, several work \citep{hsu2009multi,tai2012multilabel,chen2012feature,yu2014large} have been proposed to tackle the problem by reducing the dimensionality of label spaces in which label co-occurrence patterns play a crucial role.
In language modeling, another area of research where high dimensionality issues arise more severely, one can deal with the problem by representing words as dense vectors in $\mathbb{R}^d$ and then to learn these representations in a way of maximizing a conditional probability of a word given the context of the word \citep{bengio2003neural}.
Recently, for learning word representations, \citet{mikolove2013distributed} have proposed a very efficient log-linear model relying on co-occurrence patterns of words given the fixed-length contexts and have shown interesting properties of word representations by analogical reasoning.
In the following Section, we introduce a novel approach to reduce the dimensionality of label spaces by adapting the log-linear models capable of handling the large-scale problems efficiently.

\section{The Log-linear Model} \label{sec:loglinear_model}

As stated in Section \ref{sec:intro}, there are several work making use of hierarchical structures in label spaces for improving predictive performance of multi-label classifiers.
This is because label hierarchies are often built and maintained by human experts, and can be used as an additional source of information to identify internal structures.
Thus, we propose a log-linear model to reduce the dimensionality of label spaces by exploiting label co-occurrences and hierarchical relations between labels.

Formally, the basic idea is to maximize a probability of predicting an ancestor given a particular label in a hierarchy as well as predicting co-occurring labels with it.
Given the labels of $N$ training instances $\mathcal{D}_{\mathcal{Y}} = \{\mathcal{Y}_1, \mathcal{Y}_2, \cdots, \mathcal{Y}_{N}\}$, the objective function $\mathcal{L}$ is to maximize the average log probability given by
\begin{equation}
\mathcal{L}\left(\Theta;\mathcal{D}_{\mathcal{Y}}\right) = \sum_{n=1}^{N}\left[\frac{1}{\mathcal{Z_A}}\sum_{i\in \mathcal{Y}_n} \sum_{j \in \mathcal{S_{A}}(i)} \log p\left(y_{j} | y_{i}\right) + 
\frac{1}{\mathcal{Z_N}}\sum_{i\in \mathcal{Y}_{n}}\sum_{\substack{k \in \mathcal{Y}_{n}\\ k \neq i}}  \log p\left(y_{k} | y_{i}\right)\right]
\label{eq:label_pretrain_logprob}
\end{equation}
where ${\tiny \mathcal{Z_A} = |\mathcal{Y}_{n}||\mathcal{S}_{\mathbf{A}}(\cdot)|}$ and ${\tiny\mathcal{Z_N} = |\mathcal{Y}_{n}|(|\mathcal{Y}_{n}|-1)}$, and $\Theta$ is a set of parameters (to be introduced shortly).
The probability of predicting an ancestor label $j$ of a label $i$, i.e., $p(y_j | y_i)$ in Equation \ref{eq:label_pretrain_logprob}, can be computed by \emph{softmax} as follows
\begin{equation}
p(y_j | y_i) = \frac{\exp({\mathbf{u}'}^{T}_j\mathbf{u}_i)}{\sum_{l \in L} \exp({\mathbf{u}'}^{T}_{l}\mathbf{u}_i)}
\label{eq:softmax}
\end{equation}
where ${\mathbf{u}'}_{j}$ is a $j$-th column vector of $\mathbf{U}' \in \mathbb{R}^{d \times L}$ and ${\mathbf{u}}_{i}$ is a $i$-th column vector of $\mathbf{U} \in \mathbb{R}^{d \times L}$. 
Here, $\mathbf{U}$ and $\mathbf{U}^{\prime}$ correspond to continuous vector representations for input labels and output labels, respectively, and they are used as parameters of Equation \ref{eq:label_pretrain_logprob}, that is,  $\Theta = \{\mathbf{U}, \mathbf{U}^{\prime}\}$.
This enables for labels that share the same ancestors or co-occurring labels to have similar representations.
The softmax function in Equation \ref{eq:softmax} can be viewed as an objective function of a neural network consisting of a linear activation function in the hidden layer and two weights $\{\mathbf{U},\mathbf{U}'\}$, where $\mathbf{U}$ connects the input layer to the hidden layer while $\mathbf{U}'$ is used to convey the hidden activations to the output layer.
Thus, the log of a probability of predicting a label $j$ given a label $i$ in Eq \ref{eq:softmax}, i.e., $e_{j|i} \triangleq \log p(y_j | y_i) $, can be obtained in a neural network framework:
\begin{align}
\mathbf{h}_i &= \mathbf{U}\mathbf{y}_i \\
o_{j|i} &= {\mathbf{U}'}_{\cdot j}^{T}\mathbf{h}_i\\
e_{j|i} &= o_{j|i} -\log \sum_k \exp(o_{k|i})
\end{align} 
where $\mathbf{y}_i$ is a $L$ dimensional vector whose $i$-th element is set to one and zero elsewhere and $\mathbf{h}_i$ is hidden activation of a label $i$.
Note that since only $i$-th element of $\mathbf{y}_i$ is one, $\mathbf{h}_i$ is equal to $\mathbf{u}_i$ in Eq \ref{eq:softmax}.
The probability of predicting a label $k$, which is assigned together with a label $i$ to a label set $\mathcal{Y}_n$, $p(y_k | y_i)$ is computed in the same way meaning that it is parameterized by $\mathbf{U}$ and $\mathbf{U}^{\prime}$ as well.
The parameters are updated as follows
$
\mathbf{U}_{(t+1)} = \mathbf{U}_{(t)} + \alpha\nabla_{\mathbf{U}_{(t)}}\mathcal{L},\,
{\mathbf{U}^{\prime}}_{(t+1)} = {\mathbf{U}^{\prime}}_{(t)} + \alpha\nabla_{{\mathbf{U}^{\prime}}_{(t)}}\mathcal{L}\,
$
where $\alpha$ denotes a learning rate.

Due to computational cost for $\nabla_{{\mathbf{U}^{\prime}}_{(t)}}\mathcal{L}$, we use \emph{hierarchical softmax} \citep{morin2005hierarchical,mnih2009scalable} \ which reduces the gradient computing cost from $\mathcal{O}(L)$ to $\mathcal{O}(\log L)$.
Similar to \citet{mikolove2013distributed}, in order to make use of the \emph{hierarchical softmax}, a binary tree is constructed by Huffman coding giving binary codes with variable length to each label according to $|\mathcal{S}_{\mathbf{D}}(\cdot)|$.
In other words, the more descendants a label has in a hierarchy, the shorter codes it is assigned.
Note that by definition of the Huffman coding all labels correspond to leaf nodes in the binary tree, namely Huffman tree, and there are $L-1$ internal nodes in the Huffman tree.
Instead of computing $L$ outputs, the \emph{hierarchical softmax} computes a probability of $\lceil \log L \rceil$ binary decisions over a path from the root node of the tree to the leaf node corresponding to a target label, say, $y_j$ in Equation \ref{eq:softmax}.

More specifically, let $C(y)$ be a codeword of a label $y$ by the Huffman coding, where each bit can be either 0 or 1, and $I(C(y))$ be the number of bits in the codeword for that label.
$C_{l}(y)$ is the $l$-th bit in $y$'s codeword.
Unlike the \emph{softmax}, in computing the \emph{hierarchical softmax} we use the output label representations $\mathbf{U}^{\prime}$ as vector representations for inner nodes in the Huffman tree.  
The \emph{hierarchical softmax} is given by
\begin{equation}
p(y_j | y_i) = \prod_{l=1}^{I(C(y_j))} \sigma(\llbracket C_{l}(y_j) = 0\rrbracket \thinspace {\mathbf{u}^{\prime}}_{n(l,y_j)}^{T}\mathbf{u}_{i})
\label{eq:hierarchical_softmax}
\end{equation}
where $\sigma(\cdot)$ is the logistic function, $\llbracket \cdot \rrbracket$ denotes a function taking 1 if its argument is true, otherwise -1, and ${\mathbf{u}^{\prime}}_{n(l,y_j)}$ is a vector representation for the $l$-th node in the path from the root node to the node corresponding to the label $y_j$ in the Huffman tree.
While $L$ inner products are required to compute the normalization term in Equation \ref{eq:softmax}, the \emph{hierarchical softmax} needs $I(C(\cdot))$ times computations.
Hence, the \emph{hierarchical softmax} allows substantial improvements in computing gradients if $\mathbb{E} \left[I(C(\cdot))\right] \ll L$.


\section{Experimental Setup} \label{sec:exp_setup}

We carried out all experiments on the BioASQ Task 2a dataset in which 12 millions of training documents and approximately 27,000 index terms are given with parent-child pairs defined over index terms.\footnote{\url{http://www.bioasq.org}}
The index terms are known as Medical Subject Headings (MeSH), which is controlled and updated continually by National Library of Medicine (NLM), and used to index articles for the MEDLINE and PubMed databases.

{\bf Representing parent-child pairs in a DAG. }
It is often the case that if we represent parent-child pairs of labels as a graph, in real-world problems, it contains cycles.
Sometimes, it is reasonable to assume that these cycles result from wrong annotations due to complex relationship between labels with different abstract levels.
Additionally, cycles in a graph also introduce other difficulties to learning algorithms.
For instance, if we want to visit all ancestors given a node, where the node has a direct edge to one of its ancestors, such graph traversal never ends unless stopping criteria are used.
Hence, we remove edges resulting in cycles as follows:
\vspace{-2mm}
\begin{itemize}
\setlength{\itemsep}{1pt}
\setlength{\parskip}{0pt}
\setlength{\parsep}{0pt}
\item Pick a node that has no parent as a starting node.
\item Run Depth-First Search (DFS) from the starting node in order to detect edges pointing to nodes visited already, then remove such edges.
\item Repeat the above steps until all nodes having no parent are visited.
\end{itemize}
\vspace{-2mm}
The above preprocessing steps left us 46,455 parent-child pairs by removing 3,461 ones, from which we obtain a DAG-structured label hierarchy.

{\bf Training details.}
We set the dimension of vector representations $d$ to 128 and the learning rate $\alpha$ is fixed to $0.1$ during training.
It took about 3 hours to iterate 50 times over 12 millions of training instances on Intel Xeon E5-2620 equipped with 24 threads.

\section{Experimental Results} \label{sec:exp_results}

\begin{figure}[t]
\centering
\includegraphics[width=0.8\textwidth]{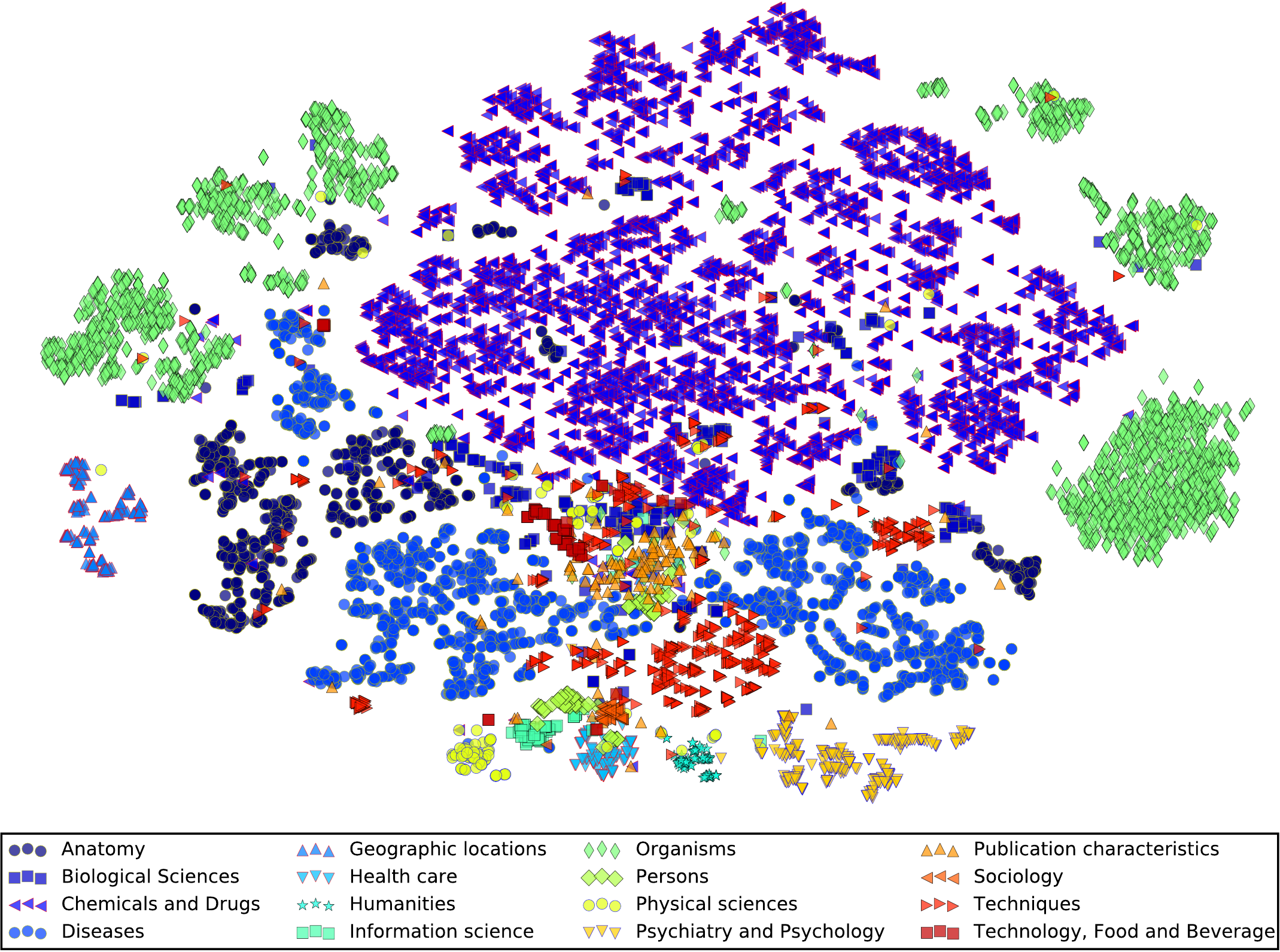}
\caption{Learned representations of 16 major categories in MeSH vocabulary. Projection of label representations into 2D is done by \emph{tSNE} \citep{van2008visualizing}.}
\label{fig:16_major_categories}
\end{figure}

The hierarchy of MeSH vocabulary consists of 16 major categories.
Each index term belongs to at least one major category.
After training the log-linear model on the BioASQ dataset, we selected labels corresponding terminals and belonging to a single major category in order to visualize learned representations.
Figure \ref{fig:16_major_categories} shows that using a hierarchy and co-occurrences of labels the model separates reasonably well 16 major categories defined in MeSH vocabulary.
Please note that whereas we use \emph{tSNE} for better visualization in Figure \ref{fig:16_major_categories}, \emph{Principal Component Analysis} is used to project label representations for the rest of figures in this paper.
In the following section, we investigate what label representations learn from a hierarchy and co-occurrences.

\subsection{Encoding Hierarchical Structures} \label{sec:enc_hierarchy}

\begin{figure}
    \centering
    \begin{subfigure}[b]{0.45\textwidth}
        {\includegraphics[width=\textwidth]{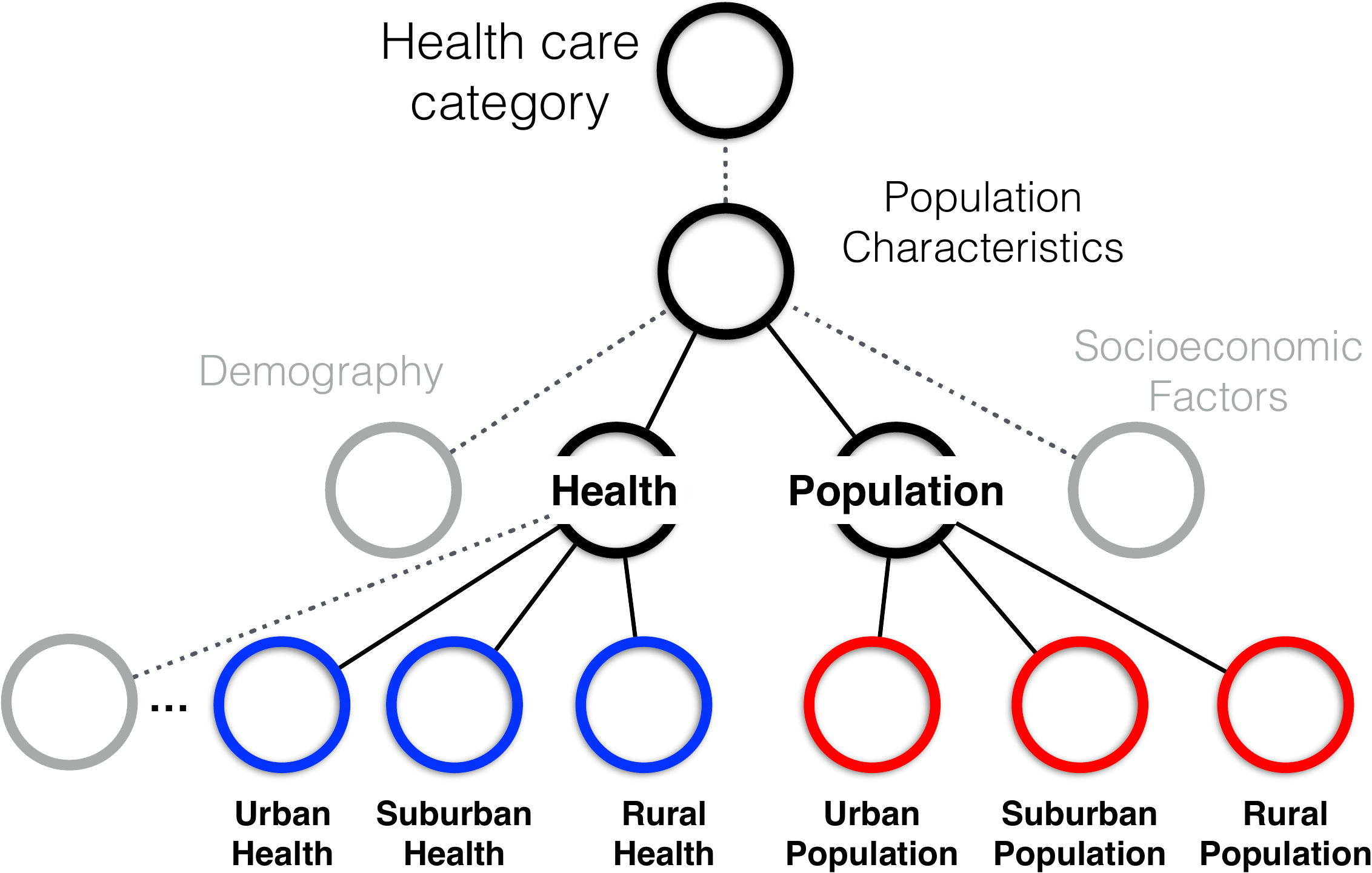}
        \subcaption{}\label{fig:orig_hierarchy}}
    \end{subfigure}
    \\
    \begin{subfigure}[b]{0.45\textwidth}
        {\includegraphics[width=\textwidth]{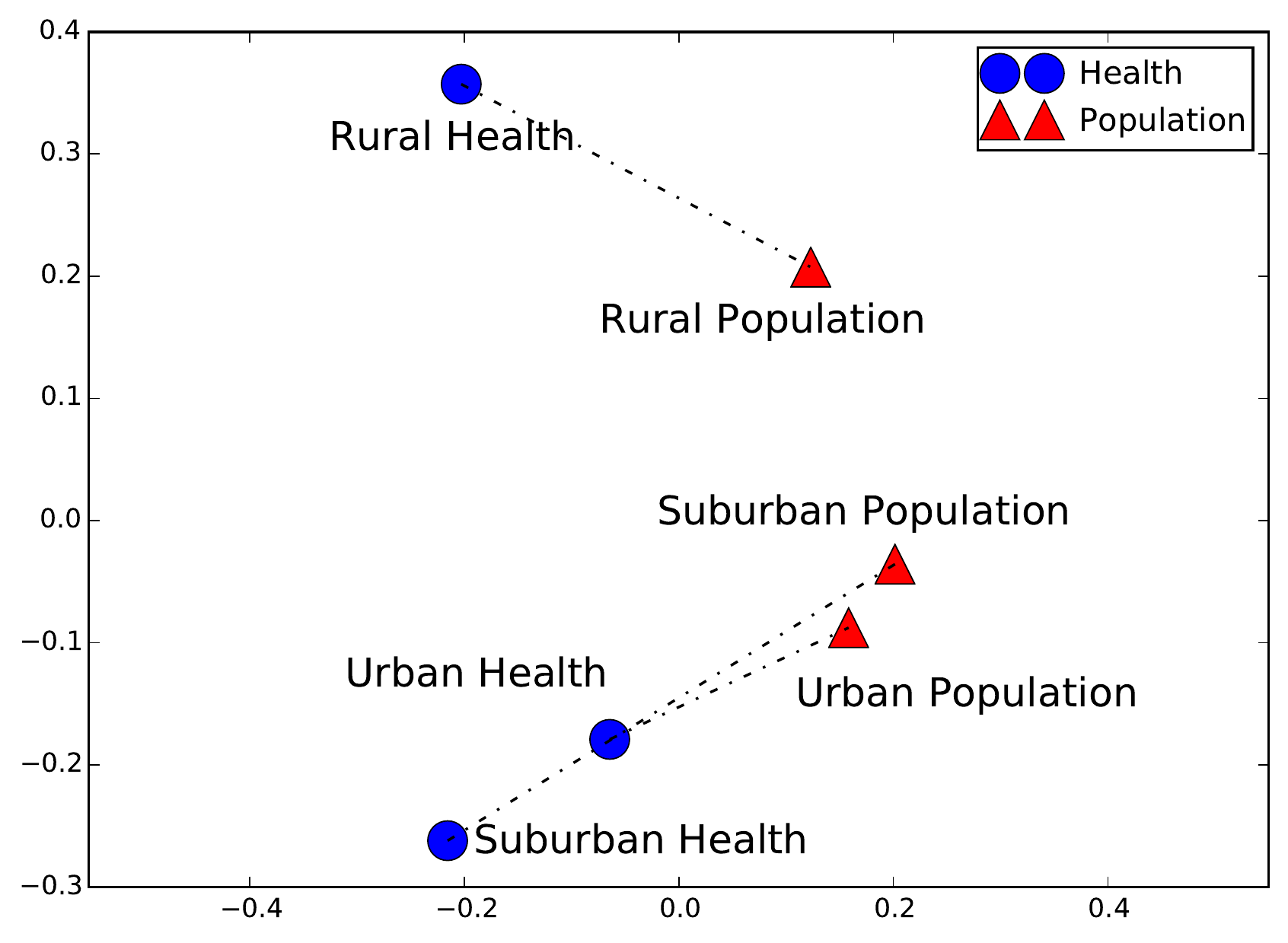}
        \subcaption{}\label{fig:health_population_occ}}    
    \end{subfigure}
    \hfill
    \begin{subfigure}[b]{0.45\textwidth}
        {\includegraphics[width=\textwidth]{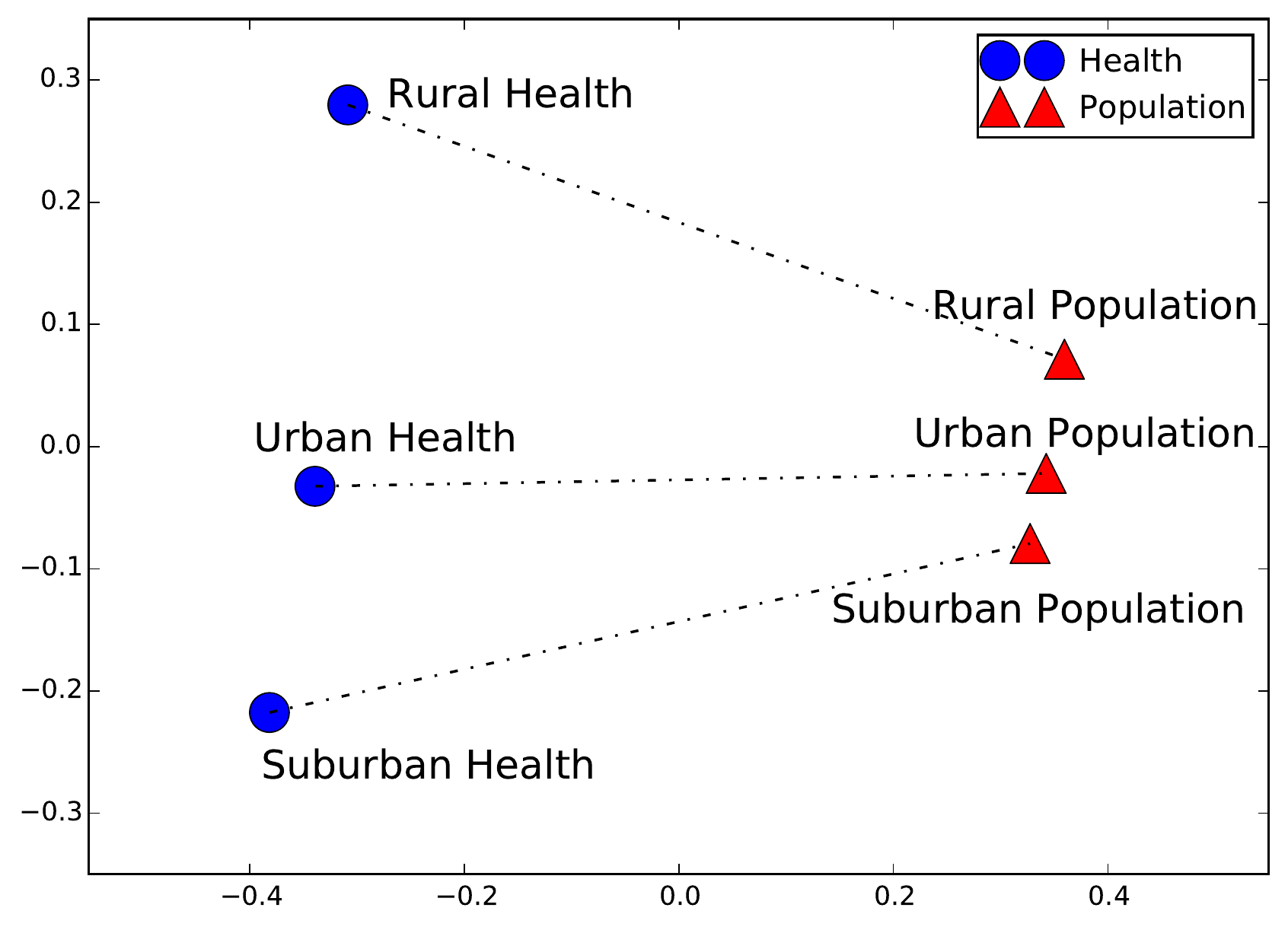}
        \subcaption{}\label{fig:health_population_hier}}
    \end{subfigure}
    \caption{\subref{fig:orig_hierarchy} A part of the hierarchy related to health care. \subref{fig:health_population_occ} Learned vector representations for the index terms at leaf \textbf{without} the hierarchy (manually rotated). \subref{fig:health_population_hier} Learned vector representations for the same index terms \textbf{with} the hierarchy.}
    \label{fig:label_struct_emb}
\end{figure}

As an illustration of our results, we will focus on a subgraph related to health care, shown in the top of Figure \ref{fig:label_struct_emb}.
Consider the leaf nodes in the figure.
According to our objective in Equation \ref{eq:label_pretrain_logprob}, \emph{Urban Health}, \emph{Suburban Health} and \emph{Rural Health} are trained to have representations for predicting their common ancestors, i.e., \emph{Health}, \emph{Population Characteristics} and \emph{Health care category}.
The child nodes of \emph{Population} are also trained similarly.
Although \emph{Urban}, \emph{Suburban} and \emph{Rural Health} are separated from \emph{Urban}, \emph{Suburban} and \emph{Rural Population}, their representations tend to be similar since they share the same ancestors.
In other words, \emph{Urban Health} and \emph{Rural Population}, for example, should have somewhat similar representations in part so as to increase a probability of predicting their common ancestors even though they are rarely assigned to the same instance.


We did perform an experiment to see whether learning from only the co-occurrences yields meaningful structures.
In this case, our objective function is limited to the right term in Equation \ref{eq:label_pretrain_logprob}.
Co-occurrence information enables to learn internal structures (see Figure \ref{fig:health_population_occ}).
If we consider hierarchical relationship between labels as well as co-occurrences of them, a more interesting property of our proposed method can be observed.
A major difference between Figure \ref{fig:health_population_occ} and \ref{fig:health_population_hier} are the inter- and intra-group relationships among labels at the leaves.
Specifically, all child nodes of \emph{Health} are located in the left side and those of \emph{Population} appear in the opposite side (Figure \ref{fig:health_population_hier}).
In addition to such relationship between two groups, relations between labels belonging to \emph{Health} (on the left) resemble those found among the children of \emph{Population} (on the right).

\begin{figure}
\centering
    \begin{subfigure}[b]{0.45\textwidth}
        \includegraphics[width=\textwidth]{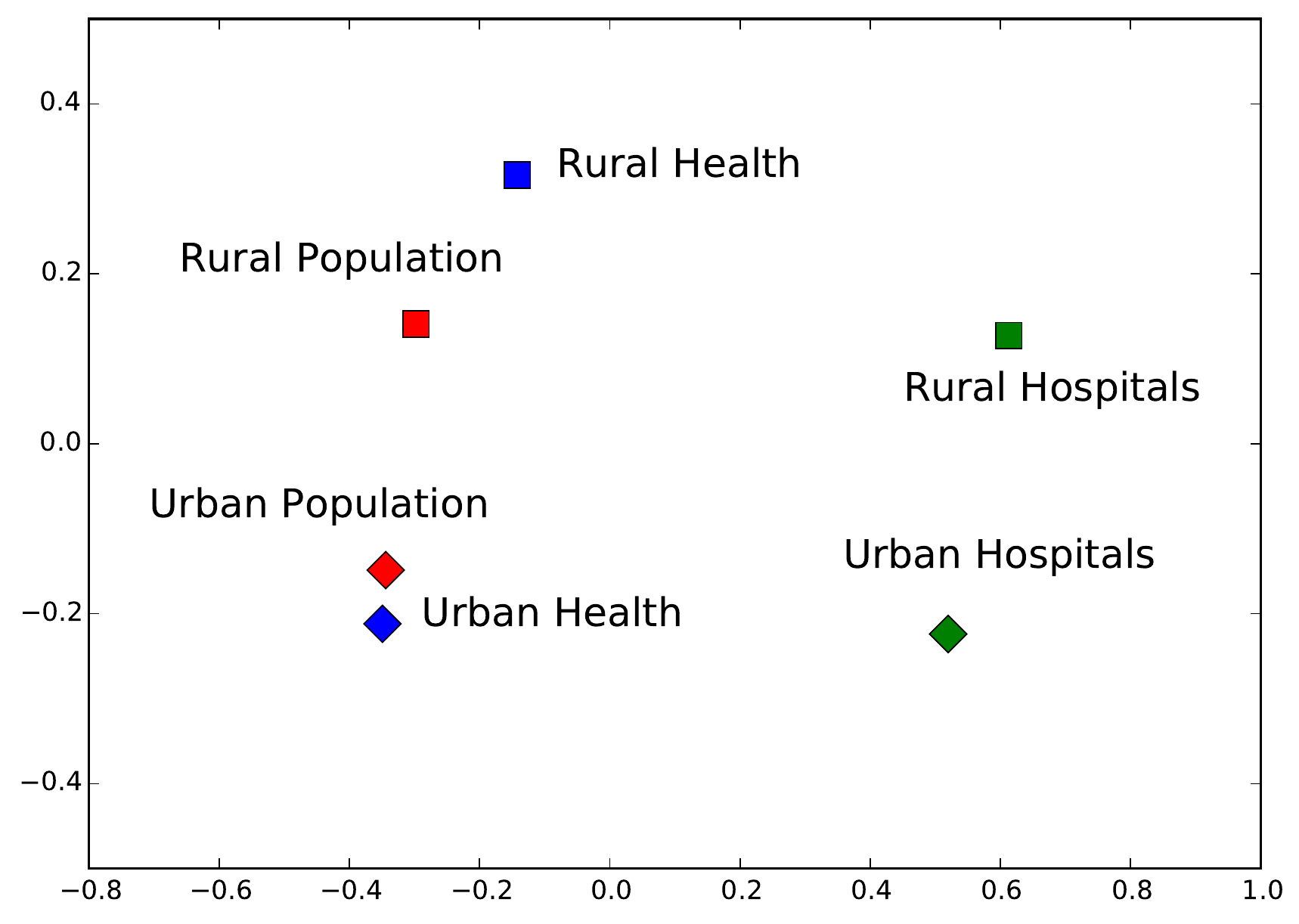}
        \subcaption{}\label{fig:urban_suburban_rural_occ}            
    \end{subfigure}
    \hfill
    \begin{subfigure}[b]{0.45\textwidth}
        \includegraphics[width=\textwidth]{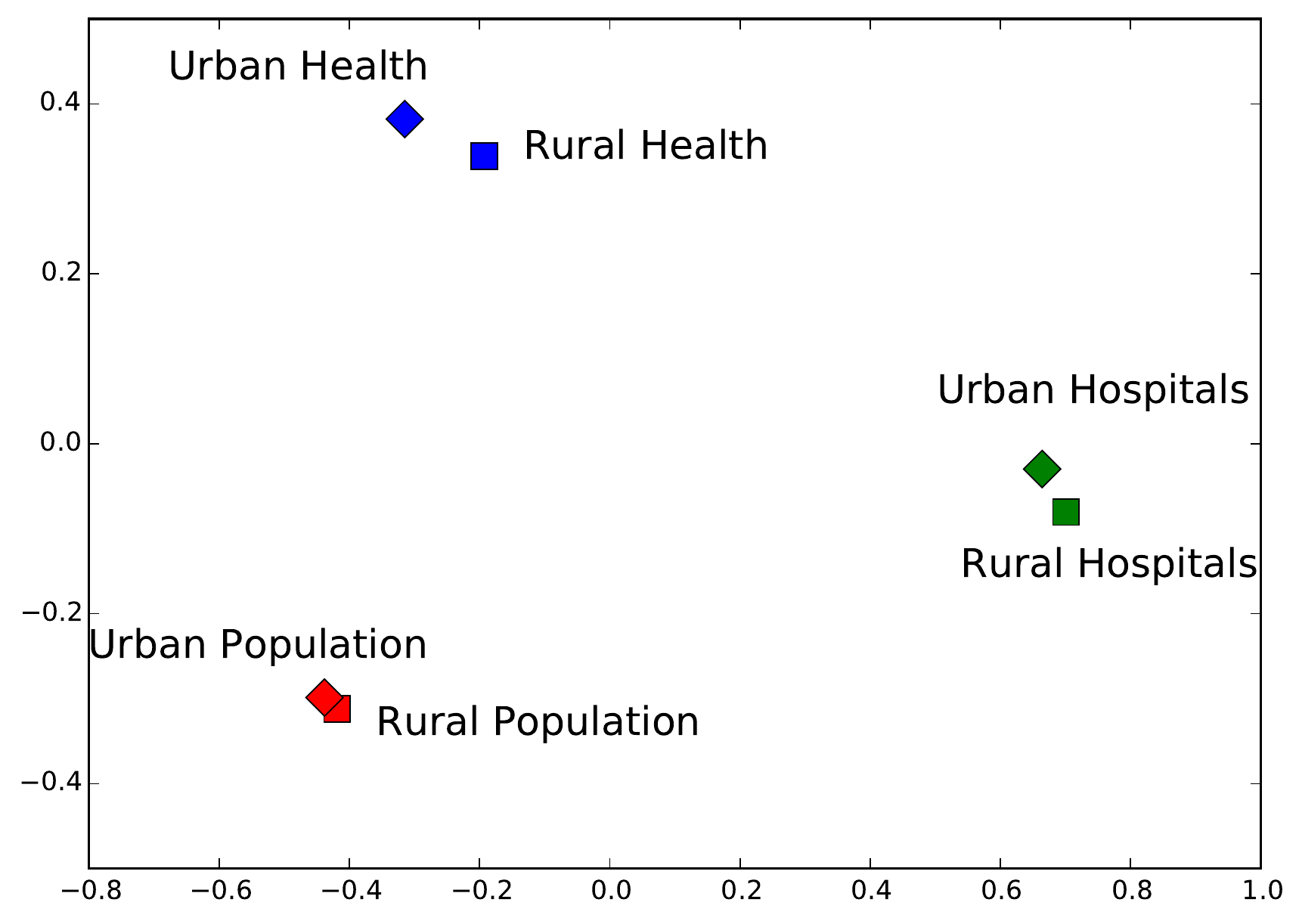}  
        \subcaption{}\label{fig:urban_suburban_rural_hier}
    \end{subfigure}\\
        \begin{subfigure}[b]{0.45\textwidth}
        \includegraphics[width=\textwidth]{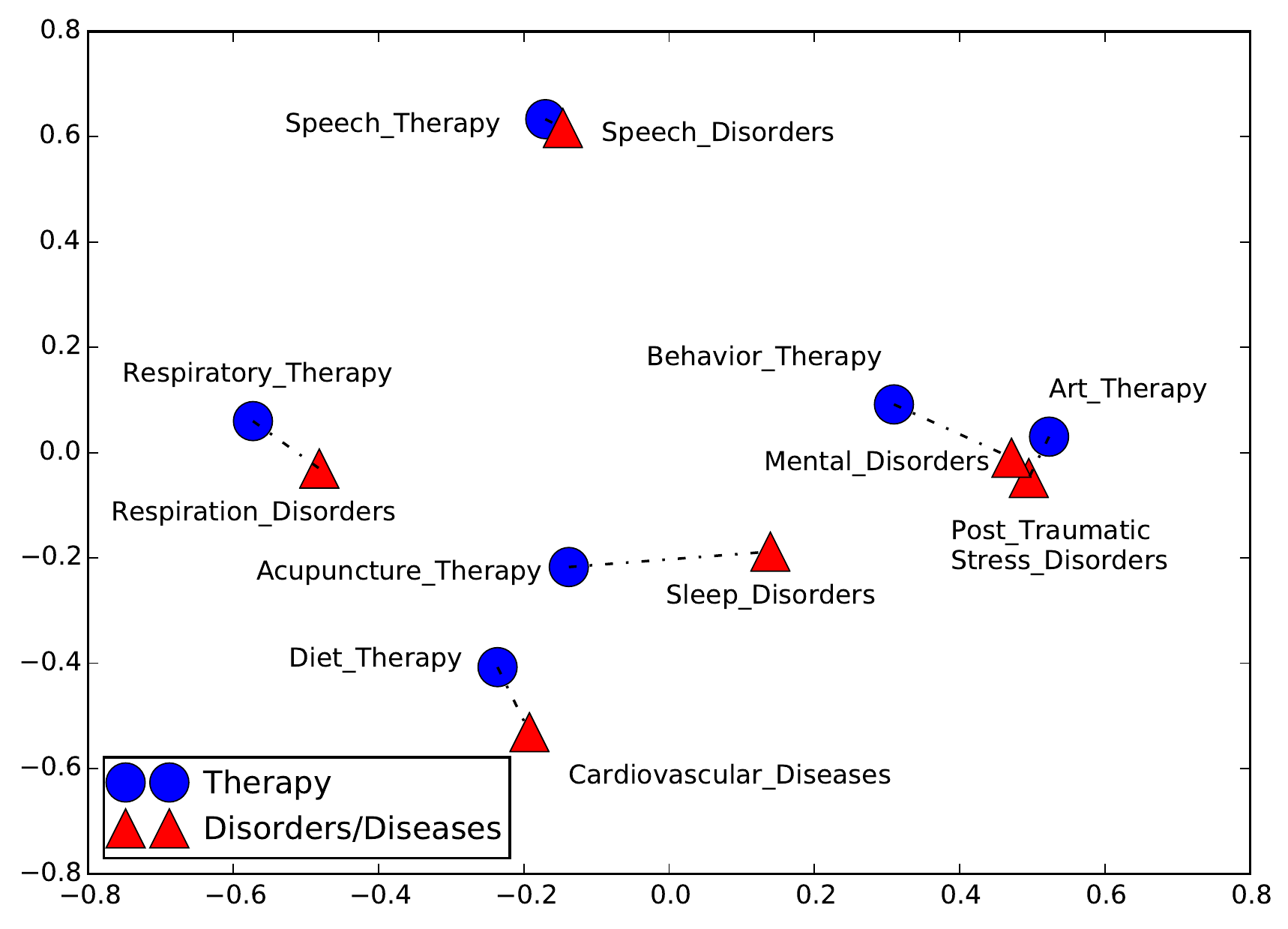}            
        \subcaption{}\label{fig:therapy_disorders_occ}
    \end{subfigure}
    \hfill
    \begin{subfigure}[b]{0.45\textwidth}
        \includegraphics[width=\textwidth]{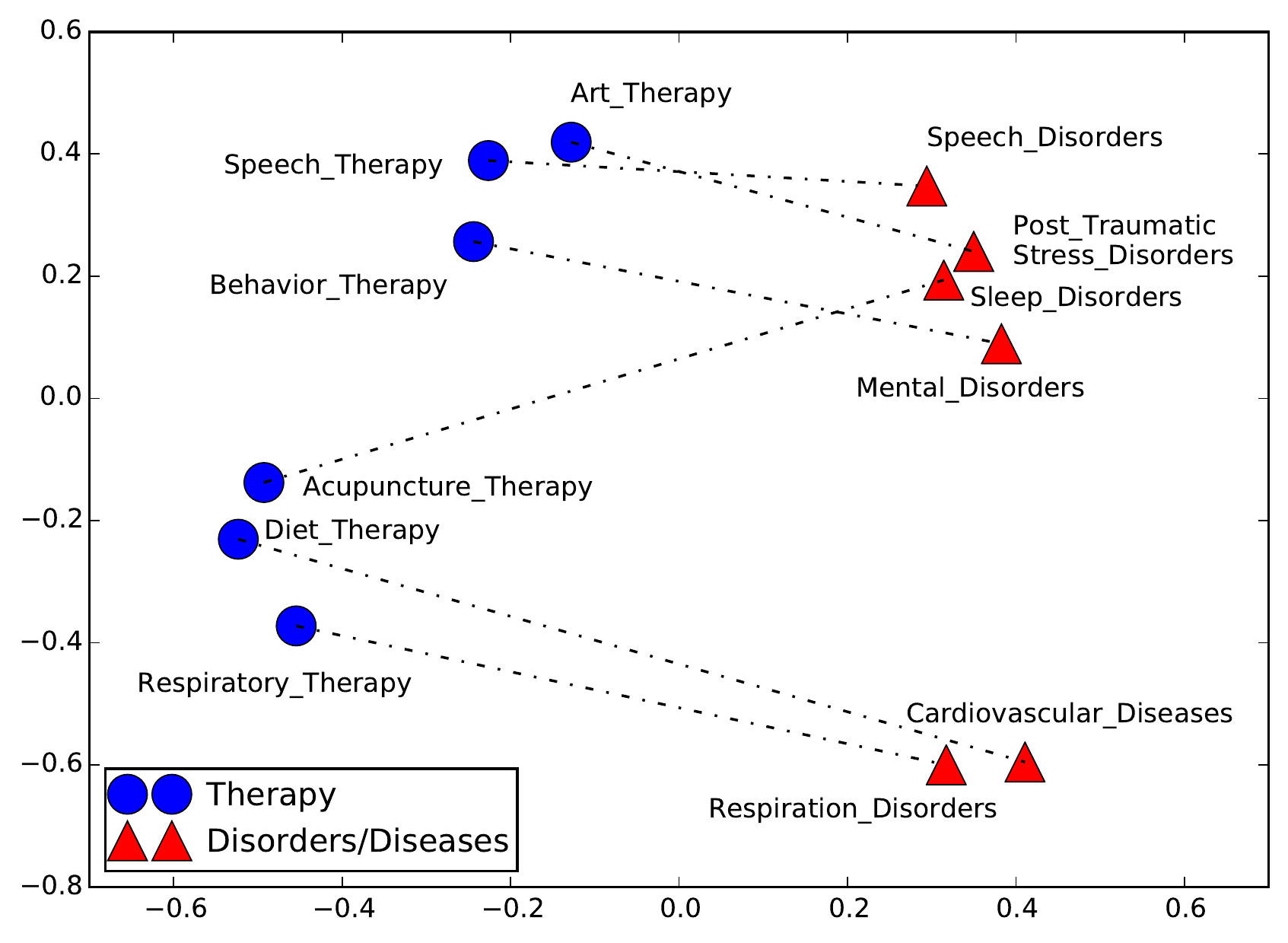}
        \subcaption{}\label{fig:therapy_disorders_hier}
    \end{subfigure}
\caption{\subref{fig:urban_suburban_rural_occ} \& \subref{fig:therapy_disorders_occ} labels' representations learned from only co-occurrences \subref{fig:urban_suburban_rural_hier} \& \subref{fig:therapy_disorders_hier} from a hierarchy as well as co-occurrences}
\label{fig:comp_occ_hierarchy}
\end{figure}

Figure \ref{fig:comp_occ_hierarchy} shows how important it is to use hierarchical information for capturing regularities of the labels.
It is likely that label pairs that co-occur frequently are close to each other (see Figure \ref{fig:urban_suburban_rural_occ} and \ref{fig:therapy_disorders_occ}).
In particular, Figure \ref{fig:urban_suburban_rural_occ} illustrates that \emph{Urban Population} is close to \emph{Urban Health} and \emph{Rural Population} because \emph{Urban Population} is often assigned together with \emph{Urban Population} and \emph{Rural Population} to a document.
Likewise, as shown in Figure \ref{fig:therapy_disorders_occ}, each therapy is close by a disorder for which the therapy is an effective treatment.
If we make use of hierarchical information during training the model, it reveals more interesting relationship which is not observed from the model trained on only co-occurrences.
Figure \ref{fig:urban_suburban_rural_hier} shows that there is a strong directed relation from \emph{Urban} to \emph{Rural} which can be represented by a direction vector pointing upper-left.
We can say the learned vector representations has identified \emph{Therapy}-\emph{Disorders/Diseases} relationship (Figure \ref{fig:therapy_disorders_hier}).

\subsection{Analogical Reasoning}

One way to evaluate representation quality is analogical reasoning as shown in \citep{mikolove2013distributed}. For example, one could represent the analogy that a king is to a queen like a man to a woman with a qualitative equation such as ``\emph{king} - \emph{man} + \emph{woman} $\approx$ \emph{queen}.''
Upon the observations in Section \ref{sec:enc_hierarchy}, we performed analogical reasoning on both the representations trained with the hierarchy and ones without the hierarchy, specifically, regarding \emph{Therapy}-\emph{Disorders/Diseases} relationship (Table \ref{tab:analogical_reasoning}).
As expected, it seems like the label representations trained with the hierarchy are clearly advantageous to ones trained without the hierarchy on analogical reasoning.
To be more specific, consider the first example, where we want to know what kinds of therapies are effective on ``Respiration Disorders'' as the relationship between ``Diet Therapy'' and ``Cardiovascular Diseases.''
When we perform such analogical reasoning using learned representations with the hierarchy, the most probable answers to this analogy question are therapies that can used to treat ``Respiration Disorders'' including nutritional therapy.
Unlike the learned representations with the hierarchy, it is likely that the label representations learned without the hierarchy perform poorly on this type of tasks.

However, we could not observe such regularities from analogical reasoning questions in terms of parent-child relationships.
For instance, it was expected that either ``\emph{Urban Health} - \emph{Health} + \emph{Population}'' or ``\emph{Urban Health} - \emph{Health} -\emph{Population Characteristics} + \emph{Population}'' results in representations close to \emph{Urban Population}, but the answers of such questions did not end up with something close to \emph{Urban Population}.
We conjecture that this is because our proposed method only encodes one-way hierarchical dependence of labels.
In other words, each index term is trained to predict its ancestors, but not to predict its descendants.

\begin{table}
\caption{Analogical reasoning on learned vector representations of MeSH vocabulary}
\label{tab:analogical_reasoning}
\begin{center}
\begin{tabular}{| c | c |}
    \hline
    Analogy questions & Most probable answers \\\hline \hline
    \multicolumn{2}{|c|}{On learned representations \emph{using} the hierarchy} \\\hline
    \pbox{20cm}{Cardiovascular Diseases : Diet Therapy \\ $\approx$ Respiration Disorders : ?} & \pbox{20cm}{\relax\ifvmode\centering\fi Diet Therapy \\ Enteral Nutrition \\ Gastrointestinal Intubation \\ Total Parenteral Nutrition \\ Parenteral Nutrition \\ Respiratory Therapy } \\\hline
    \pbox{20cm}{Mental Disorders : Behavior Therapy \\ $\approx$ Post Traumatic Stress Disorders : ?} & \pbox{20cm}{\relax\ifvmode\centering\fi Behavior Therapy \\ Cognitive Therapy \\ Rational-Emotive Psychotherapy \\ Brief Psychotherapy \\ Psychologic Desensitization \\ Implosive Therapy}\\\hline\hline
    \multicolumn{2}{|c|}{On learned representations \emph{without using} the hierarchy} \\\hline    
    \pbox{20cm}{Cardiovascular Diseases : Diet Therapy \\ $\approx$ Respiration Disorders : ?} & \pbox{20cm}{\relax\ifvmode\centering\fi Respiration Disorders \\ Respiratory Tract Diseases \\ Respiratory Sounds \\ Airway Obstruction \\ Hypoventilation \\ Croup } \\\hline
    \pbox{20cm}{Mental Disorders : Behavior Therapy \\ $\approx$ Post Traumatic Stress Disorders : ?} & \pbox{20cm}{\relax\ifvmode\centering\fi Behavior Therapy \\ Psychologic Desensitization \\ Internal-External Control \\ Post Traumatic Stress Disorders \\ Phobic Disorders \\ Anger}\\\hline
\end{tabular}
\end{center}
\end{table}

\subsection{Different Hierarchies, Different Representations}

\begin{figure}
    \centering
    \begin{subfigure}[b]{\textwidth}
        \includegraphics[width=\textwidth]{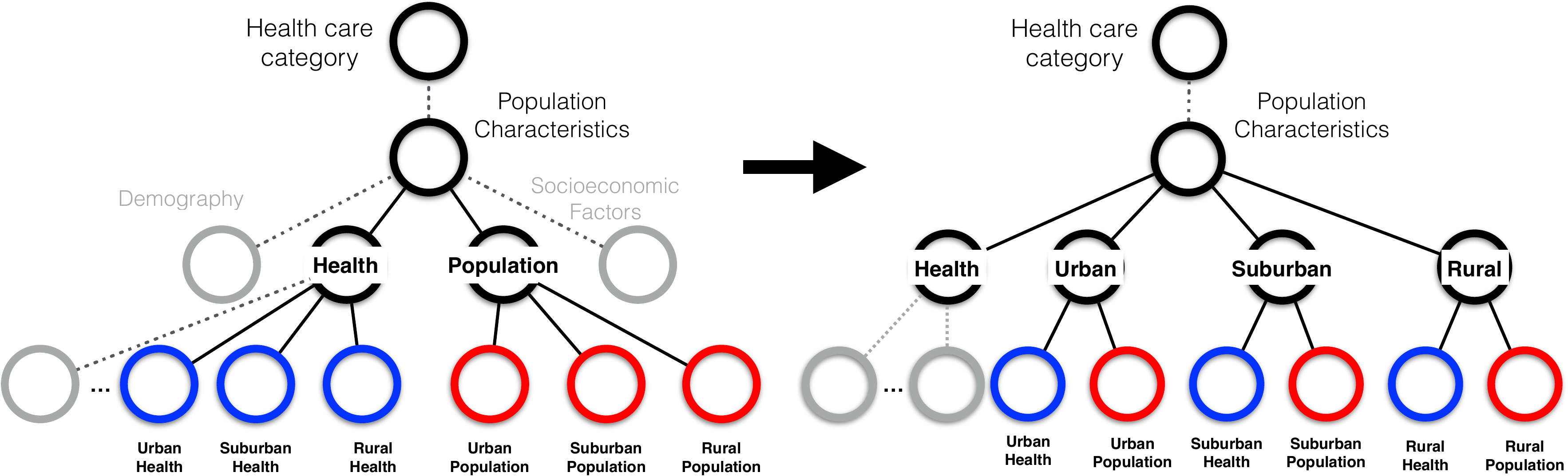}
        \subcaption{}\label{fig:modified_hier}
    \end{subfigure}
    \\
    \begin{subfigure}[b]{0.32\textwidth}
        {\includegraphics[width=\textwidth]{figures/health_population_without_hierarchy.pdf}
        \subcaption{}\label{fig:again_health_population_occ}}    
    \end{subfigure}
    \begin{subfigure}[b]{0.32\textwidth}
        {\includegraphics[width=\textwidth]{figures/health_population_with_hierarchy.pdf}
        \subcaption{}\label{fig:again_health_population_hier}}
    \end{subfigure}
    \begin{subfigure}[b]{0.32\textwidth}
        \includegraphics[width=\textwidth]{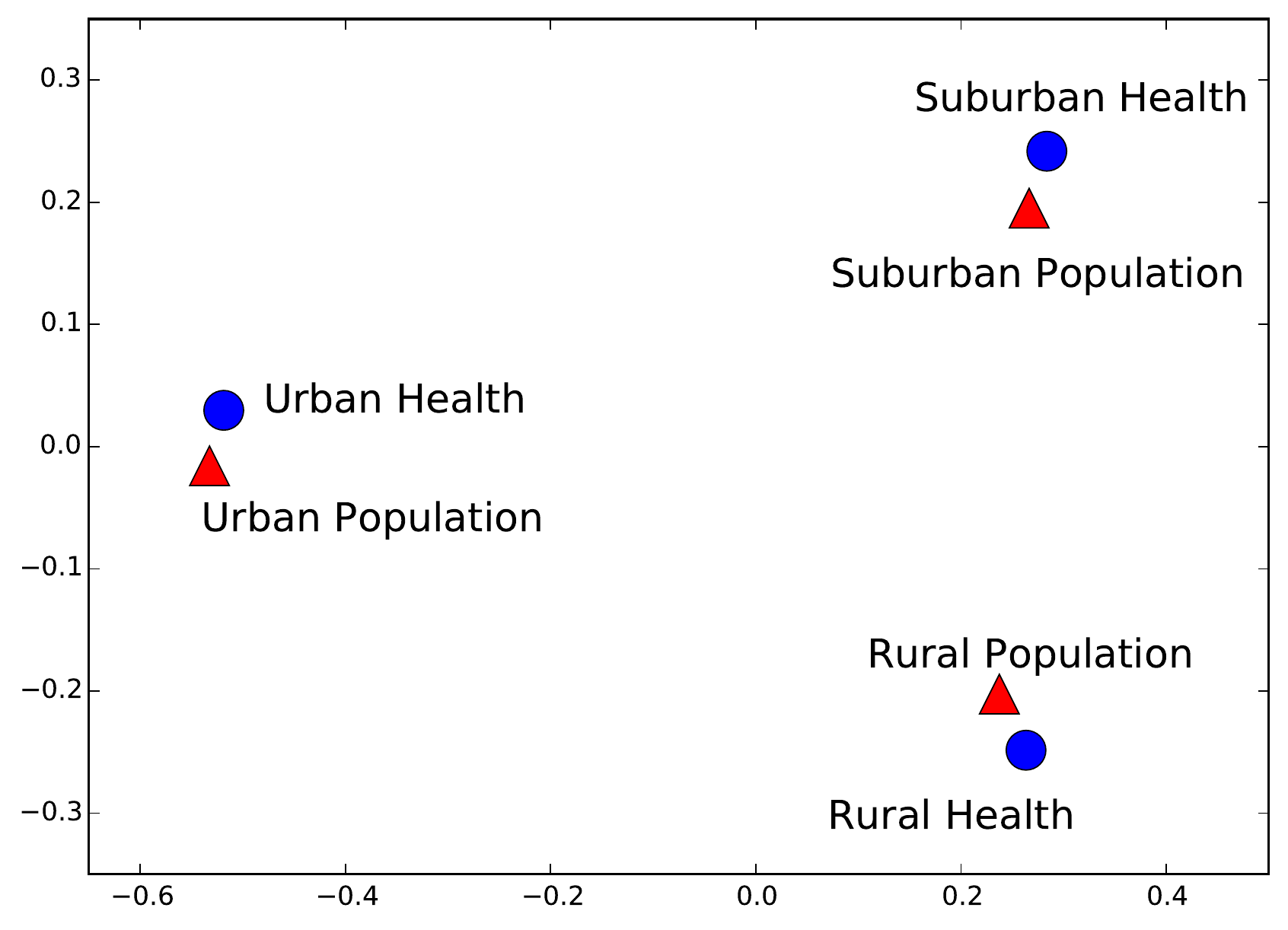}
        \subcaption{}\label{fig:modified_hier_emb}
    \end{subfigure}
    \caption{\subref{fig:modified_hier} A modified hierarchy (\emph{right}) from the original one (\emph{left}) obtained by grouping the same types of developed environments. See text for further explanation.
    \subref{fig:again_health_population_occ} Learned vector representations without using a hierarchy.
    \subref{fig:again_health_population_hier} Learned vector representations using the \emph{original} hierarchy. \subref{fig:modified_hier_emb} Learned vector representations using the \emph{modified} hierarchy.}
    \label{fig:modified_hier_and_emb}
\end{figure}

In the previous section, we have shown our proposed method is capable of capturing hierarchical structures over labels and co-occurrences of them.
In this case, it can be expected that the learned representations change when some part of the hierarchy is changed while the label co-occurrences remains same.
To answer this question, firstly, we modified the original hierarchy (left in Figure \ref{fig:modified_hier}) so as to obtain the new hierarchy (right in Figure \ref{fig:modified_hier}).
Originally, \emph{Population Characteristics} has two child nodes, \emph{Health} and \emph{Population}.
Contrary to \emph{Population} that has only three child nodes, \emph{Health} has dozens of child nodes.
Instead of removing the other nodes from the hierarchy, we kept them as they are in the hierarchy.
However, \emph{Population} was removed from the hierarchy.
We then created three new internal nodes, \emph{Urban}, \emph{Suburban} and \emph{Rural} representing types of developed environments.
Finally, all nodes of interest (in blue and red) were grouped according to the developed types.

Figure \ref{fig:modified_hier_emb} shows learned vector representations with modified hierarchy and compares the learned representations with the previous results.\footnote{For  the sake of readability, we used repeated results, the left graph in \subref{fig:modified_hier}, \subref{fig:again_health_population_occ} and \subref{fig:again_health_population_hier} taken from Figure \ref{fig:label_struct_emb}, in order to clearly show the difference between representations learned from the original hierarchy and from the modified one as well as learned from only co-occurrences.}
Unlike the previous result in Figure \ref{fig:again_health_population_hier}, where \emph{Health}-related nodes and \emph{Population}-related nodes form clusters, \emph{Urban Health} and \emph{Urban Population} are clustered together since they share the common parent node, \emph{Urban}. 
We can also observe similar patterns for \emph{Suburban} and \emph{Rural}.
Besides, relative distance between each \emph{Population} and \emph{Health} within a cluster is identical, and the same direction vector from \emph{Health} to \emph{Population} or the other way around can be defined.
Please note that, in this case, learning the model only from co-occurrences yields always similar label representations since we only modified the hierarchy (Figure \ref{fig:again_health_population_occ}).

\section{Discussion and Future Work} \label{sec:discussion}
We have presented a method that learns vector representation of labels taking hierarchical structures into account.
The empirical results demonstrate that using hierarchical structures of labels with label co-occurrences have more impact on identifying regularities of labels than using label co-occurrences only.

We evaluated label representations qualitatively by observing label representations in 2D.
Even though inter- and intra-group relations can be found in the learned label space, it is still desired to evaluate our method using quantitative measures because we have compared and analyzed both learned representations with the very limited number of examples and chosen arbitrarily.
Hence, we are currently attempting to evaluate the model quantitatively to check whether the model brings on better prediction results in multi-label classification as a means of pre-training a joint embedding space \citep{weston2010large}.
We are also interested in extending the model to be capable of analogical reasoning on parent-child relationship.

%

\bibliography{iclr2015}
\bibliographystyle{iclr2015}

\end{document}